\pdfoutput=1

\documentclass[11pt]{article}

\usepackage{ACL2023}

\usepackage{times}
\usepackage{latexsym}
\usepackage{multirow}
\usepackage{graphicx}
\usepackage{amsfonts}
\usepackage{bbm}

\usepackage[T1]{fontenc}

\usepackage[utf8]{inputenc}

\usepackage{microtype}

\usepackage{inconsolata}

\title{Sebis at SemEval-2023 Task 7: A Joint System for Natural Language Inference and Evidence Retrieval from Clinical Trial Reports}

\author{Juraj Vladika \and Florian Matthes \\
  Department of Computer Science, Technical University of Munich \\
  \texttt{ \{juraj.vladika, matthes\}@tum.de} \\}

\begin{document}
\maketitle
\begin{abstract}
With the increasing number of clinical trial reports generated every day, it is becoming hard to keep up with novel discoveries that inform evidence-based healthcare recommendations. To help automate this process and assist medical experts, NLP solutions are being developed. This motivated the SemEval-2023 Task 7, where the goal was to develop an NLP system for two tasks: evidence retrieval and natural language inference from clinical trial data. In this paper, we describe our two developed systems. The first one is a pipeline system that models the two tasks separately, while the second one is a joint system that learns the two tasks simultaneously with a shared representation and a multi-task learning approach. The final system combines their outputs in an ensemble system. We formalize the models, present their characteristics and challenges, and provide an analysis of achieved results. Our system ranked 3rd out of 40 participants with a final submission. 
\end{abstract}

\section{Introduction}

Clinical trials are research studies carried out in human subjects to assess the effectiveness and safety of medical, surgical, or behavioral interventions \cite{friedman2015fundamentals}. These investigations constitute the main approach that medical researchers use to determine whether a novel treatment such as a new medication, procedure, diet, or medical device is safe and efficient in humans. Clinical trials are often used to compare the effectiveness of a new treatment against the standard treatment or placebo treatment and to assess its adverse effects. When performed rigorously, clinical trials present the most valuable resource for informing evidence-based healthcare decisions. 

\begin{figure}[h]
    \centering
    \includegraphics[width=0.5\textwidth]{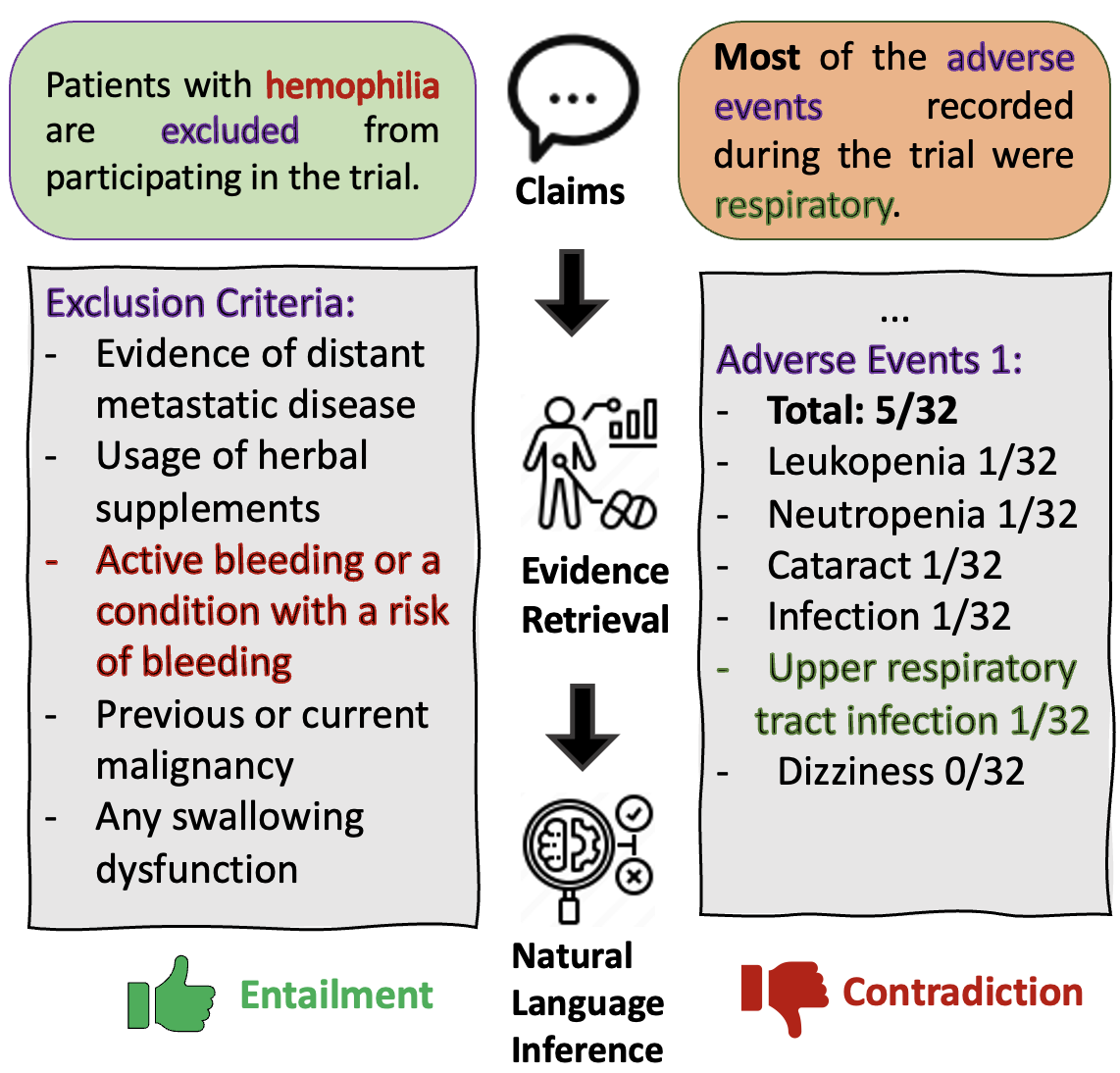}
    \caption{The task consists of predicting whether a given claim entails or contradicts the clinical trial report based on the evidence found in it.}
    \label{fig:task}
\end{figure}

On average, more than 100 reports of CTs are published every day \cite{zarin201910}. Keeping up with all their results and novel discoveries is a time-consuming and often practically impossible endeavor. This has brought to light dedicated organizations, such as Cochrane,\footnote{\url{https://www.cochranelibrary.com}} that manually synthesize these clinical outcomes \cite{higgins2019cochrane}. Nevertheless, they struggle to keep up with the ever-increasing amount of literature and new studies. For this purpose, automated approaches based on Machine Learning (ML) and Natural Language Processing (NLP) can be developed to facilitate the process of inferring new knowledge and finding evidence from clinical trial reports.

The SemEval-2023 Task 7, titled \textit{Multi-evidence Natural Language Inference for Clinical Trial Data (NLI4CT)}, focused on developing NLP systems for making conclusions and finding evidence in clinical trial reports (CTRs) \cite{jullien-2023-nli4ct}. It featured two subtasks that were strongly coupled together. The first subtask was, given a clinical trial document and a claim, to develop a model that infers a logical relation between them, namely either \textsc{Entailment} or \textsc{Contradiction}. The second subtask was, given all the sentences in the clinical trial document, to develop a model that selects a subset of those sentences that serve as evidence for making a decision on the logical entailment relation between the claim and the document. The task is illustrated in Figure \ref{fig:task}. It shows two different claims, each coupled with an excerpt from a clinical trial report and a gold final label. The gold evidence sentences are highlighted for emphasis.

For this task, we developed two competitive NLP systems. The first one is a pipeline system, which learns to perform the tasks of evidence retrieval and textual entailment separately and takes into account one sentence at a time during the evidence selection process. The second approach is a joint system, which jointly learns the two subtasks with a unified representation of a claim and the whole document, and uses a shared training procedure. The joint system performed better on the test set, but the final system is an ensemble system that utilizes the power of both systems by combining their outputs with an averaging function. Our final system achieved the 3rd place in the competition out of 40 submitting participants. We provide the code of our system in a GitHub repository.\footnote{\url{https://github.com/jvladika/NLI4CT}}

\section{Related Work}
The task of Natural Language Inference (NLI) consists of inferring whether there is logical entailment or contradiction between two pieces of text -- a premise and a hypothesis. It has been researched since the early days of NLP when it was mostly tackled with rule-based and linguistically informed approaches \cite{maccartney2009natural}. A big surge of interest in the task occurred after the release of the SNLI dataset \cite{bowman-etal-2015-large}, as well as follow-up datasets like MultiNLI (MNLI) \cite{williams-etal-2018-broad} and Adversarial-NLI (ANLI) \cite{nie-etal-2020-adversarial}. The task has also been researched in the clinical domain -- the dataset MedNLI \cite{romanov-shivade-2018-lessons} features more than 14,000 pairs of claims from clinical notes and patient records, extracted from the MIMIC-III database \cite{johnson2016mimic}.

The task of evidence retrieval, or more precisely evidence sentence selection, has been studied as a step in common NLP tasks like Machine Reading Comprehension (MRC) \cite{wang-etal-2019-evidence} and Question Answering (QA) \cite{thayaparan-etal-2021-textgraphs}. Its main purpose is to improve the explainability and interpretability of models' predictions but also to aid its reasoning process for making decisions. Combining the steps of evidence retrieval and textual entailment recognition comprises the task of automated fact-checking or claim verification, where the goal is to assess a claim's veracity based on the evidence related to it \cite{guo-etal-2022-survey}. This task has mostly been concerned with verifying claims related to news events, society, and politics, but has been increasingly so researched in scientific, biomedical, and clinical domains \cite{wadden-etal-2020-fact, sarrouti-etal-2021-evidence-based}.

\section{Dataset}
The clinical trials used for constructing the dataset originate from \textit{ClinicalTrials.gov},\footnote{\url{https://clinicaltrials.gov}} a database of privately and publicly funded clinical studies conducted around the world, run 
by the U.S. National Library of Medicine. All the clinical trial reports in the dataset are related to breast cancer and are written in English. There is a total of 1,200 clinical trials and 2,400 corresponding claims related to these CTRs, which were written and annotated by clinical domain experts, clinical trial organizers, and research oncologists. 

For the task, each CTR may contain 1-2 patient groups, called cohorts or arms, and these groups may receive different treatments, or have different baseline characteristics. The dataset consists of a total of 2,400 claims and was split into a training dataset of 1,700 claims, a development dataset of 200 claims, and a hidden test dataset of 500 claims. There are two types of claims, those related only to a single CTR and other ones related to two CTRs, which usually make some form of comparison between the two reports. Each of the claims is related to only one of the following four sections:

\begin{itemize}
    \item \textbf{Eligibility Criteria}. A set of conditions for patients to be allowed to take part in the clinical trial, such as age, gender, medical history.
    \item \textbf{Intervention}: Information concerning the type, dosage, frequency, and duration of treatments being studied in the clinical trial.
    \item \textbf{Results}: Reports the outcome of the trial with data like the number of participants, outcome measures, units, and results.
    \item \textbf{Adverse Events}: These are any unwanted side effects, signs, or symptoms observed in patients during the clinical trial.
\end{itemize}

\section{System Description}

In this chapter, we will describe the architecture of the developed systems and the choice of base models comprising the systems. Even though the task features two standalone subtasks and participating in each of them was optional, we found the best synergy is achieved by combining them. The selection of appropriate evidence sentences in subtask 2 is an important prerequisite for recognizing textual entailment in subtask 1. This is because most textual entailment datasets used for training the NLI systems are modeled to recognize entailment between two sentences only, so the performance between a sentence and a whole document is still underwhelming \cite{schuster-etal-2022-stretching}. Therefore, it was important to narrow down the whole document to a shorter span of text with only the relevant evidence sentences.

\begin{table*}[t]
\centering

\begin{tabular}{l|ccc|ccc}

\hline
\multicolumn{1}{l}{}                & \multicolumn{3}{c}{\textbf{Evidence Retrieval}} & \multicolumn{3}{c}{\textbf{Textual Entailment}} \\ \hline
 \textbf{Base Model}  & \textbf{Precision} & \textbf{Recall   }            & \textbf{F1 }                       & \textbf{Precision} & \textbf{Recall   }            & \textbf{F1  }  \\ \hline
 BERT & $77.5 $ & $80.8 $ & $80.2 $ & $58.2 $ & $64.0 $ & $61.0 $  \\  \hline 
BioBERT & $78.0 $ & $81.9 $ & $80.8 $ & $64.3 $ & $65.0 $ & $64.5 $  \\ 
 ClinicalBERT & $81.0  $    & $81.2 $   & $81.1  $  & $52.7  $  & $87.0 $ & $65.7  $        \\ 
UmlsBERT & $78.6 $ & $88.5 $ & $83.9 $ & $56.1 $ & $83.0 $ & $67.0 $  \\ \hline

StructBERT & $77.1 $ & $86.5 $ & $82.4 $ & $50.3 $ & $91.0 $ & $64.8 $ \\ 
ERNIE & $ 84.5  $    & $81.2  $   & $84.8  $  & $53.0  $  & $88.0  $ & $66.2  $        \\ 
DeBERTa-v3 & $81.8 $ & $89.1 $ & $\textbf{86.2 }$ & $78.5 $ & $84.0 $ & $\textbf{80.5} $   \\ \hline

\end{tabular}

\caption{\label{tab:results}Results of different base models for the two tasks on the development set.}
\end{table*}

\subsection{Pipeline Systems}
In the pipeline system, the evidence sentences selected by the first model are used as the input for veracity prediction to the next model. Using the standard terminology from computer science, we call these systems pipeline systems. It is also common to use the same underlying base model in both steps and fine-tune it for these two different tasks \cite{deyoung-etal-2020-eraser}.

To formalize the approach, we will define it mathematically. Given a claim $c$ and $n$ sentences $s_1, s_2, ...,  s_n$ that constitute the clinical trial report, the goal is to train a model that predicts $$z_i = \mathbbm{1} [s_i \: \textrm{is an evidence sentence}].$$ This is modeled as a binary sequence classification task, where candidate sequences are a concatenation of a candidate sentence $s_i$ and the claim $c$ in the form of $a_i = [s_i; SEP; c]$. Each of these sequences is encoded with the base language model to obtain their dense representation $h_i = BERT(a_i)$. This representation is then fed to a classifier model, Multi-Layer Perceptron (MLP), and its output is passed to a softmax function that assigns the probabilities on whether the candidate sentence is or is not the evidence: 
$$p_i, \, \bar{p}_i = softmax(MLP(h_i)).$$ Finally, a selector function labels those sentences with a probability over a threshold to be evidence sentences -- this threshold is a hyperparameter that can be learned, such as $z_i = p_i > 0.5$. In the end, the model has selected $k$ final evidence sentences $e_1, e_2, ..., e_k$ that are used as input for the next step.

The task of textual entailment is again a binary classification task that for a given claim $c$ and $k$ evidence sentences $e_1, e_2, ..., e_k$ predicts the logical relation of either \textsc{Entailment} or \textsc{Contradiction}. Using the standard terminology of Natural Language Inference, the claim $c$ is the hypothesis, and the concatenation of evidence sentences $e = [e_1; e_2; ...; e_k]$ is the premise. These two are concatenated as $x = [c; SEP; e]$ and once again passed to the base language model to obtain the dense embedding $w = BERT(x)$. The model has to learn the function $$\hat{y}(c; e) = softmax(MLP(w)),$$ which is the probability distribution of each inference label for the claim $c$ given evidence $e$. The class with the highest probability score is selected as the final verdict $v(c; e) = argmax(y)$.

\subsection{Joint Systems}
The second system we developed jointly learns both the tasks of evidence retrieval and textual entailment. This leverages the machine-learning technique of multi-task learning (MTL), which was shown to be data efficient and improve the performance of each individual task it learns on through shared representations \cite{Crawshaw2020MultiTaskLW}. Intuitively, the model improves the performance on both tasks simultaneously since selecting high-quality evidence is important for recognizing entailment and conversely, the final entailment/contradiction label influences the specific evidence to be selected.

Unlike in the pipeline system, where each sequence consisted only of \textit{one} candidate sentence and the claim, here the claim $c$ is concatenated together with \textit{all} of the sentences $s_1, s_2, ..., s_n$ in the clinical trial document to obtain a claim-document sequence $seq = [c; SEP; s_1; SEP;$ $s_2; ...; SEP; s_n]$.\footnote{Considering the input sequence length limitation of large language models, this sequence has to be truncated to $512$ or $1024$ tokens, which is good enough for the vast majority of documents in the dataset.} This approach makes the representation of each candidate sentence aware of the context it appears in with regard to the rest of the document, as well as aware of the claim itself. This sequence is fed to the base language model to obtain a dense representation $h = BERT(seq) = [h_c; SEP; h_{s_1}; ...: SEP; h_{s_n}]$. The representation of each candidate sentence $h_{s_i} = [h_{w_1}, h_{w_2}, ..., h_{w_m}]$ is singled out from the initial representation and passed to a binary linear classifier that, similarly to the one in the pipeline system, calculates the probabilities of the sentence being an evidence sentence: 
$$p_i, \, \bar{p}_i = softmax(MLP(h_{s_i})).$$ Those sentences that are above the $0.5$ threshold are then selected as evidence sentences and concatenated together to form the final evidence representation $h_e = [h_{e_1}, h_{e_2}, ..., h_{e_k}]$. This representation is given to a final ternary linear classifier, that same as in the pipeline system predicts the verdict $$v = argmax(softmax(MLP(h_e))).$$ Note that the representation of the claim $h_c$ is not passed anywhere because the idea is for evidence sentences to already be aware of the semantics of the claim from the joint claim-document representation.

\subsection{Ensemble System}
After developing the pipeline system and the joint system, we ended up with the best-performing systems on the two tasks in each category of systems. These best systems were singled out and their outputs were combined for the final system. This type of approach is usually called a stacked \cite{pavlyshenko2018using} or an ensemble system \cite{ganaie2021ensemble} and has been proven to perform well in machine-learning shared tasks and competitions. 

Considering that the outputs of the two systems are different in certain predictions because of different output probabilities of classes, we decided to average the probabilities of each of the systems with appropriate weights. After experimenting with different weights, the final function was: $$p_{final} = 0.4 \cdot p_{pipeline} + 0.6 \cdot p_{joint}.$$

\subsection{Base Models}
Both of the components constituting the system use an underlying base model. We opt for large pre-trained language models (PLMs) since they represent the state of the art in virtually all NLP tasks. We experimented with a number of different base models. BERT \cite{devlin-etal-2019-bert} is used as the representative vanilla PLM, which gives good initial insight into the performance of PLMs on the task. Over the years, there have been multiple domain-specific variations of the BERT model, specialized for text in the scientific, biomedical, or clinical domains. BioBERT \cite{lee2020biobert} was fine-tuned on abstracts of biomedical scientific publications. ClinicalBERT \cite{alsentzer2019publicly} is a model fine-tuned on MIMIC-III database, a database of electronic health records and clinical notes of patients admitted to critical care units. UmlsBERT \cite{michalopoulos2020umlsbert} moves further away from unstructured text and injects structured domain knowledge, from UMLS \cite{bodenreider2004unified} -- a large knowledge base of biomedical concepts and semantic relations between them, into the model training process. 

DeBERTa \cite{he2021deberta} is an improved extension of the BERT model that introduced disentangled attention and enhanced masked decoder, which both amplify the importance of positional embedding of tokens in a sequence. The DeBERTa-v3 \cite{DBLP:journals/corr/abs-2111-09543} is a novel version of the model that uses the task of replaced token detection (RTD) instead of predicting masked tokens. It was the first model to beat the human performance on the GLUE benchmark \cite{wang2018glue} for natural language understanding tasks. For comparison, we include other models that excel in NLU tasks, namely StructBERT \cite{DBLP:conf/iclr/0225BYWXBPS20} and ERNIE \cite{zhang-etal-2019-ernie}. In the end, we would like to test whether the models highly specialized for inference and textual entailment tasks beat the BERT extension models fine-tuned to biomedical language and knowledge.

The hyperparameters were the same for all models and setups: learning rate $10^{-5}$, warmup rate $0.06$, weight decay $0.01$, epochs $5$--$7$, mixed precision training enabled. For all of the datasets, their \textit{Large} version was used, imported from the HuggingFace repository.\footnote{For example, DeBERTa-v3-Large is at: \url{https://huggingface.co/microsoft/deberta-v3-large}} The models were additionally fine-tuned on the previously mentioned NLI datasets like MNLI, ANLI, MedNLI, following the approach of \citet{Laurer-van-Atteveldt-Casas-Welbers-2022}.

\section{Results \& Analysis}
In this section, we report on results achieved by our systems and provide a qualitative error analysis with challenging examples from the dataset.

\subsection{Final Results}

The results of different base models for the two tasks with the pipeline system are presented in Table \ref{tab:results}. These are the results on the dev set considering that this dataset has the revealed truth labels and allowed for an unlimited number of test runs. The results for the evidence retrieval are binary classification metrics, where each candidate sentence was of positive class if it constituted evidence, otherwise of negative class. The results of the textual entailment task are again binary classification metrics for the two classes of entailment and contradiction, using the gold evidence sentences. We found that the model trained on gold evidence performs better on the final test set than the model trained on internally selected evidence sentences.

\begin{table}[htbp]
\centering
\resizebox{0.49\textwidth}{!}{
\begin{tabular}{lcc}
\hline \hline
\textbf{System}   & \textbf{Evidence F1} & \textbf{Entailment F1} \\ \hline \hline
\multirow{2}{*}{Best pipeline}    & \large \multirow{2}{*}{79.8}     & \large \multirow{2}{*}{77.2}    \\
 &    &    \\ 
\multirow{2}{*}{Best joint}   & \large \multirow{2}{*}{80.4}     & \large \multirow{2}{*}{78.3}    \\
 &    &    \\ 
\multirow{2}{*}{Best ensemble}    & \large \multirow{2}{*}{\textbf{81.8}}     & \large \multirow{2}{*}{\textbf{79.8}}    \\ 
&    &    \\ \hline
\multirow{2}{*}{Place}    & \large \multirow{2}{*}
{$4\textsuperscript{th}$}     & \large \multirow{2}{*}{$3\textsuperscript{rd}$}    \\
 &    &    \\ \hline \hline 
\end{tabular}
}
\caption{Best performing model results on the test set}
\label{tab:final}
\vspace{-5pt}
\end{table}

As visible from Table \ref{tab:results}, the NLI task of textual entailment recognition was more challenging than the evidence sentence selection task. The domain-specific biomedical and clinical BERT models outperformed the vanilla BERT model on both tasks, which shows the benefit of pre-training large language models on specialized text for NLP tasks concerning specific domains. The clinical BERT outperforms the BioBERT, which is an expected outcome since it works with clinical trial reports, while UmlsBERT outperforms both, showing the effectiveness of injecting structured knowledge into language models. Nevertheless, by far the best-performing base model turned out to be DeBERTa-v3 and it especially excelled in the textual entailment task. Although this was expected considering its proven efficiency on NLU tasks, the sheer margin of difference came as a surprise. 

The joint systems turned out to perform slightly better on the evidence retrieval task. This was expected considering that it considers the full document when doing the evidence retrieval task, which means each candidate sentence is contextualized with appropriate surroundings around it and was already shown to perform well in evidence selection for fact-verification datasets \cite{stammbach-2021-evidence}. Considering the long training times of the joint system, since it uses dense representations of full documents which can be up to $1024$ tokens, we narrowed it down to just using DeBERTa as the underlying base model for embedding generation. The scores of the best-performing pipeline and best-performing joint system on the hidden test set are shown in Table \ref{tab:final}. The outputs of these two systems were averaged with appropriate weights to obtain the final ensemble submission. We performed some additional post-processing such as truncating the selected evidence to a maximum of $20$ sentences. This was done due to the tendency of the model to select all the candidate sentences as evidence sentences for claims with quantifiers like \textit{at least one, exactly one, none of the patients}, etc. Selecting all the sentences incorrectly diminishes precision while increasing recall, but achieving high precision contributed more to the final score and was more challenging in general.

\subsection{Error Analysis}
\begin{table*}[t]
    \centering
    \begin{tabular}{ p{0.139\linewidth}   p{0.33\linewidth}  p{0.304\linewidth}  p{0.127\linewidth}  }
    \hline
      \large \textbf{Challenge} & \large \textbf{Claim}  & \large  \textbf{Evidence} & \large \textbf{Label} \\ \hline
      \textbf{Commonsense} \newline \textbf{Reasoning} & In order to participate in the trial, participants must be aware of where they are, and what day it is.         &   Inclusion Criteria:   Cognitively oriented to time, place, and person (determined by nurse)   & \color{teal} \textbf{Entailment}  \\ \hline 
        \textbf{Numerical} \newline \textbf{Reasoning} & Neutropenia affected the majority of patients in cohort 1 of the primary trial.     &   Adverse Events:  Total 26/69 (37.68\%) [...] Neutropenia 4/69 (5.80\%) & \color{red} \textbf{Contradiction}  \\ \hline 
        \textbf{Multi-Hop} \newline \textbf{Reasoning} & The primary trial and the secondary trial do not use the same route of administration for interventions.    &  \textit{(primary)} Intervention: Vaccine Therapy / \textit{(secondary)} Intervention: PET Guided Biopsy & \color{teal} \textbf{Entailment}   \\ \hline 
        \textbf{Medical} \newline \textbf{Knowledge} & Patients with cancer that has spread from a breast tumor to their CNS are able to take part in the trial.          &  No histologically proven bone marrow metastasis.   & \color{red} \textbf{Contradiction}  \\ \hline 
        \textbf{World} \newline \textbf{Knowledge} & A minimum bodyweight of 50kg is required to participate in the secondary trial.         &   Patient Characteristics:  Total body weight > 110 lbs  (without clothes)   & \color{teal} \textbf{Entailment}  \\ \hline 
      
    \end{tabular}
    \caption{Five common challenges found in the dataset with a representative example for each challenge that was classified incorrectly by our system.}
    \label{tab:challenges}
\end{table*}

To better understand the strengths and weaknesses of our system, we conducted a manual analysis of predictions on the development set to find out where the final model incorrectly labeled the claims as the opposite class. Table \ref{tab:challenges} provides five such examples from the dataset. Each example in the table consists of a claim, accompanying evidence sentences, and a gold label assigned to them by the expert annotators. We also identify five different challenges that the model has to tackle in order to be able to correctly classify these claims. These challenges are:

\begin{itemize}
  \item \textbf{Commonsense Reasoning.} The model has to understand concepts and make judgments about everyday matters, that are common and innate to humans. 
  \item \textbf{Numerical Reasoning.} The model has to be capable of applying basic mathematical operations and grasping comparisons, orders, or quantities. 
  \item \textbf{Multi-hop Reasoning.} In some examples, the model needs to combine multiple pieces of evidence (make multiple "hops") to come to the final conclusion.
  \item \textbf{Medical Knowledge.} Certain claims rely on expert knowledge of medical concepts and terminology related to the human body, diseases, drugs, or treatments.
  \item \textbf{World Knowledge.} This type of knowledge refers to the non-linguistic information about the outside world contained in claims.
\end{itemize}

\section{Conclusion}
In this paper, we describe our solution for the SemEval-2023 Task 7, dealing with natural language inference and evidence retrieval from clinical trial reports. We motivate the task, discuss related work, provide formal definitions of the developed systems, present the results, analyze the performance of models, and discuss some challenges in the process. We developed two types of systems -- a pipeline system, which learns evidence retrieval and textual entailment sequentially, and a joint system, which learns the two tasks simultaneously. The final system combines them into an ensemble and achieved the 3rd place in the competition out of 40 teams with a final submission. 

We anticipate this system will be useful for facilitating the work of medical experts in synthesizing the results and outcomes of the ever-increasing amount of clinical trial reports, as well as be used by the NLP community for the related tasks of biomedical question answering, automated claim verification, and recognizing textual entailment. The system could be improved in the future by overcoming the challenges related to commonsense, numerical, and multi-hop reasoning, or by injecting additional medical and world knowledge to it.

\bibliography{anthology,custom}
\bibliographystyle{acl_natbib}

\end{document}